\documentclass{article}

\usepackage{PRIMEarxiv}
\usepackage{amsmath}
\usepackage[utf8]{inputenc} 
\usepackage[T1]{fontenc}    
\usepackage[hypertexnames=false]{hyperref}       
\usepackage{url}            
\usepackage{booktabs}       
\usepackage{amsfonts}       
\usepackage{nicefrac}       
\usepackage{microtype}      
\usepackage{lipsum}
\usepackage{fancyhdr}       
\usepackage{graphicx}       
\usepackage{algorithm}
\usepackage{algpseudocode}
\graphicspath{{media/}}     

\title{Generative AI like ChatGPT in Blockchain Federated Learning: use cases, opportunities and future
}
\author{
  Sai Puppala, Ismail Hossain, Md Jahangir Alam, Sajedul Talukder\\
  School of Computing \\
  Southern Illinois University \\
  Carbondale, IL, USA\\
  \texttt{\{saimaniteja.puppala, ismail.hossain, mdjahangir.alam, sajedul.talukder\}@siu.edu}
\\
\And
  Jannatul Ferdaus \\
  Computer Science and Engineering \\
  University of Asia Pacific \\
  Dhaka, Bangladesh\\
\texttt{syeda.jannat243@gmail.com} 
\\
\And
  Mahedi Hasan \\
  Computer Science and Engineering \\
  Green University of Bangladesh \\
  Dhaka, Bangladesh\\
\texttt{mehedigub2020@gmail.com}
\\
\And
  Sameera Pisupati \\
  Computer Science \\
  University of North Texas \\
  TX, USA\\
\texttt{sameerapisupati@my.unt.edu}
\\
\And
   Shanmukh Mathukumilli \\
  Engineering for professionals \\
  John Hopkins University  \\
  MD, USA\\
\texttt{smathuk1@jh.edu}
}

\begin{document}
\maketitle

\begin{abstract}
Federated learning has become a significant approach for training machine learning models using decentralized data without necessitating the sharing of this data. Recently, the incorporation of generative artificial intelligence (AI) methods has provided new possibilities for improving privacy, augmenting data, and customizing models. This research explores potential integrations of generative AI in federated learning, revealing various opportunities to enhance privacy, data efficiency, and model performance. It particularly emphasizes the importance of generative models like generative adversarial networks (GANs) and variational autoencoders (VAEs) in creating synthetic data that replicates the distribution of real data. Generating synthetic data helps federated learning address challenges related to limited data availability and supports robust model development. Additionally, we examine various applications of generative AI in federated learning that enable more personalized solutions.

\end{abstract}

\section{Introduction}
In the age of big data, the emphasis has shifted from merely accumulating vast amounts of information to addressing the critical issues of data privacy and security. Data breaches have become a major concern, leading to heightened public awareness about data protection. Individuals, organizations, and society as a whole are now more focused on safeguarding data privacy and security. A prominent example of this is the General Data Protection Regulation (GDPR) introduced by the European Union. GDPR aims to protect users' personal privacy and data security by mandating clear user agreements and prohibiting misleading practices that could force users to give up their privacy rights. Additionally, operators must obtain user consent before training models and users have the right to delete their private data. When engaging in data transactions with third parties, it is essential to ensure that the contract clearly defines the scope of the data being exchanged and the data protection responsibilities. These laws and regulations pose new challenges to traditional artificial intelligence data processing methods.

In the field of artificial intelligence, data is a fundamental component necessary for model training. However, data often exists in isolated clusters, known as data islands. The conventional method to address data islands involves centralizing the data for processing, which includes collecting the data, applying standardized processing techniques, performing data cleaning, and building models. Unfortunately, data leakage often occurs during these collection and processing stages. Although regulations have strengthened the protection of users' private information, collecting data for model training has become more difficult. This has led to a significant focus on finding legal solutions to the problem of data islands within the artificial intelligence community.

\begin{figure}[h]
\centering
\includegraphics[width=0.7\textwidth]{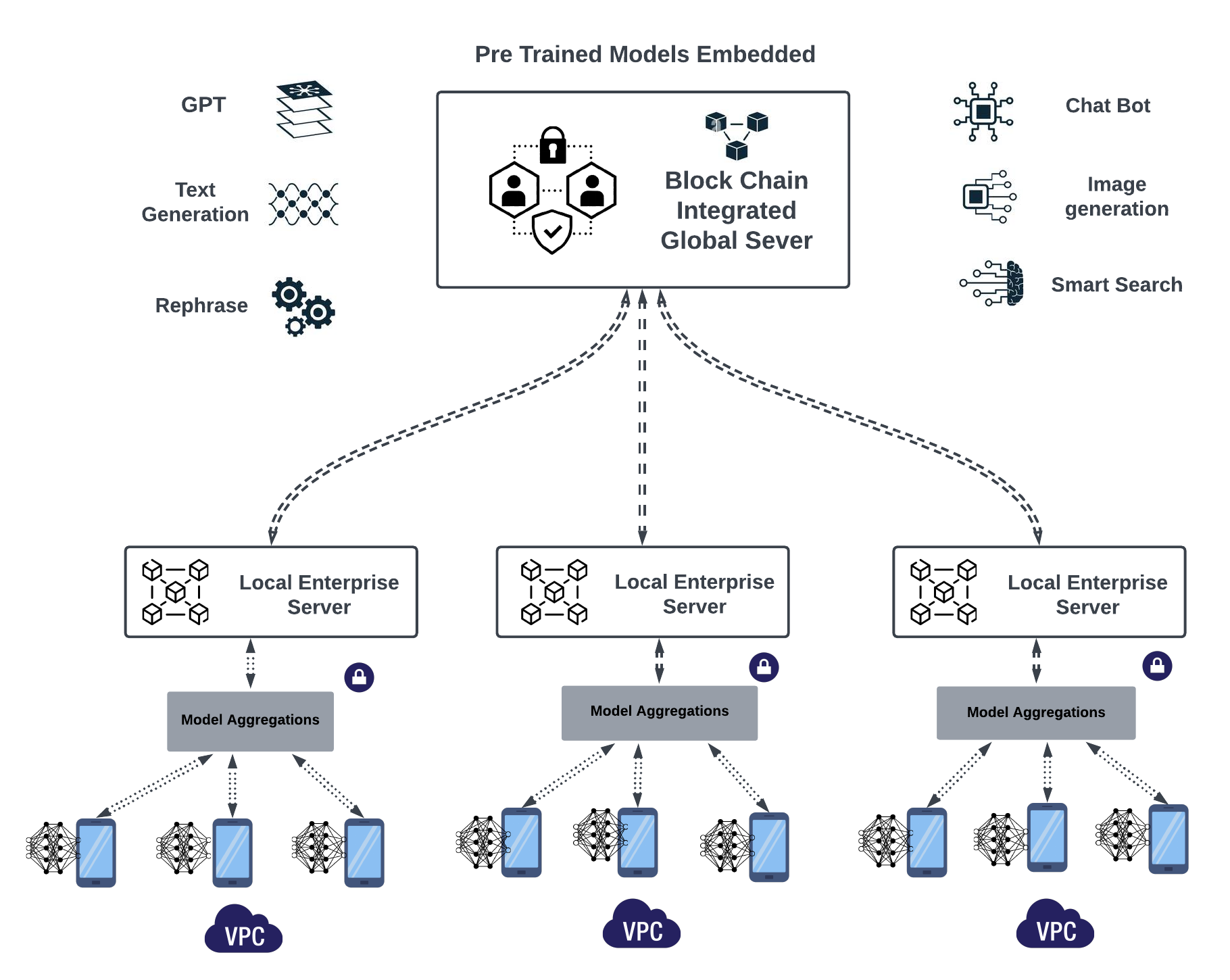}
\caption{The overview architecture of both federated learning and blockchain technology, enabling secure, efficient, and privacy-preserving distributed machine learning. } 
\label{fig:arch}
\vspace{-10pt}
\end{figure}

To address the challenges posed by data silos, traditional statistical methods have struggled to comply with various regulations, leading research to focus on solving the problem of data islands. Conventional machine learning typically relies on centralized training methods, which involve gathering training data on a single server. However, due to stringent data privacy protection regulations, implementing centralized training methods that risk data leakage and privacy breaches has become increasingly difficult. In a centralized training setup, mobile phone users who wish to train machine learning models with their data often find their individual data volume insufficient. Before the advent of federated learning, these users had to send their personal data to a central server, where machine learning models were trained using aggregated data from multiple users.

Unlike centralized training methods, federated learning is a form of distributed training. It allows users in different locations to collaboratively train machine learning models while keeping all personal data containing sensitive information on their own devices. With federated learning, users can benefit from well-trained machine learning models without having to share their privacy-sensitive data with a central server.

Federated learning has opened new research opportunities in artificial intelligence by introducing a novel training approach that enables the development of personalized models while preserving user privacy. With the advent of artificial intelligence chipsets, client devices now possess enhanced computing capabilities, facilitating the shift of artificial intelligence model training from central servers to end-user devices. Federated learning provides a privacy protection mechanism that utilizes the computing resources of end-user devices for model training, thereby preventing the leakage of private information during data transmission. Given the vast number of mobile and other field-specific devices, federated learning can fully exploit a wealth of valuable dataset resources.

Generative AI refers to the field of artificial intelligence focused on developing models and algorithms capable of creating new content resembling data from a specific domain. The core concept of generative AI is to train models to understand the underlying patterns and structures of a dataset and then generate new samples with similar characteristics. Generative models aim to capture the probability distribution of the training data and generate new instances by sampling from this learned distribution. These models can be trained in a supervised or unsupervised manner, depending on the availability of labeled data.

Generative AI in federated learning involves applying generative models within the framework of federated learning, where data remains decentralized and training occurs locally on individual devices or servers. Generative models, such as generative adversarial networks (GANs) or variational autoencoders (VAEs), can generate new samples that resemble the training data distribution.

In federated learning, the goal is to train a global model using data from multiple participants without sharing the raw data. Instead, model updates are exchanged between the participants and aggregated to improve the global model. Generative AI techniques can be incorporated into federated learning in several ways:
Lists are easy to create:
\begin{itemize}
  \item\textit{Synthetic Data Generation:} Generative models can be used to generate synthetic data that resembles the real data distribution. This synthetic data can be used to augment the limited data available at each participant, improving the robustness and generalization of the trained model.
  \item \textit{Privacy-Preserving Data Generation:} Privacy is a crucial concern in federated learning. Generative models can help address privacy concerns by generating synthetic data that preserves privacy. By training generative models on local data without sharing it, participants can generate privacy-preserving synthetic data samples that retain the statistical properties of the original data while hiding sensitive information.
  \item \textit{Data Heterogeneity:} In federated learning, the participating devices or servers may have different distributions of data. Generative models can learn the underlying data distribution of each participant and generate synthetic data that matches their respective distributions. This approach allows the global model to account for the heterogeneity of the data across different participants.
  \item \textit{Model Personalization:} Generative models can be used to personalize the global model for individual participants in federated learning. By generating personalized synthetic data samples based on a participant's local data, the global model can be fine-tuned to better cater to the specific needs or preferences of each participant.
\end{itemize}

\section{Background}
\label{sec:background}
\subsection{Federated Learning}

Federated Learning (FL) has garnered significant interest from researchers aiming to explore its capabilities and practical implementations~\cite{chen2020wireless}. Traditional centralized methods, which involve collecting and storing customer data in a central repository, are increasingly seen as impractical due to stringent data privacy regulations~\cite{geyer2017differentially}. FL is particularly advantageous when dealing with on-device data that is highly sensitive, not easily transferable to servers, or undesirable to share for various reasons~\cite{bonawitz2019towards}. In the traditional cloud server model, data from mobile devices is sent and processed centrally, raising privacy concerns among data owners and suffering from significant propagation delays and high latency~\cite{saeed2020federated}. These issues have driven the development of new techniques, resulting in the adoption of FL~\cite{lim2020federated}.

Implementing FL in various industries presents several challenges. Collaborative learning, a fundamental aspect of FL, involves training algorithms using decentralized data from different devices or servers without sharing the actual data~\cite{brik2020federated}. This approach contrasts sharply with traditional methods, which typically involve uploading data samples to servers or storing them in distributed architectures. FL emphasizes stronger modeling without data sharing, leading to more secure solutions with controlled data access privileges~\cite{li2021model}.

The main challenge is training the data without centralizing it. FL addresses this issue by focusing on collaboration, which is often infeasible with conventional machine learning techniques. Moreover, FL allows algorithms to learn from experiences accumulated across multiple devices, which is not always possible with other ML approaches~\cite{nguyen2021federated}. Rather than aggregating data from multiple sources or relying heavily on traditional discovery and replication methods, FL trains a shared global model using a central server while keeping the data in its original locations~\cite{xu2021federated}.

The architecture of an FL server must support communities of varying sizes, from tens to millions of devices, and handle rounds involving several devices up to thousands or even millions of participants~\cite{li2020review}. The updates gathered and transmitted in each round can range in size from a few kilobytes to tens of megabytes~\cite{wu2019distributed}. Additionally, device activity and charging patterns can cause significant variation in the volume of incoming and outgoing traffic in specific geographical regions throughout the day~\cite{pokhrel2020federated}.

FL performs optimally when on-device data is more pertinent than server-stored data, particularly when data privacy is paramount or when transferring data to servers is impractical or undesirable~\cite{khan2021federated}.

FL has applications in numerous fields, including healthcare, the Internet of Things (IoT), transportation, defense, education, and mobile applications~\cite{alazab2021federated, hossain2023collaborative, puppala2022towards, talukder2022novel, talukder2022federated}. The reliability of FL has been established through numerous experiments~\cite{kumar2021blockchain}. However, despite the vast potential of FL, certain technological aspects related to platforms, software, hardware, data security, and access remain not well understood~\cite{pang2020realizing}.

The increasing availability of data from various sources, including healthcare institutions, individual patients, insurance companies, and therapeutic industries, has driven significant interest in healthcare data processing in recent years~\cite{li2021model}. Figure 1 illustrates the fundamental structure of Federated Learning.

\subsection{Blockchain Federated Learning}

We can conceptualize Federated Learning (FL) clients as the blockchain nodes in a Blockchain-Federated Learning (BCFL) system. In this scenario, clients do not only train their local models but also verify updates and generate new blocks. Given this BCFL framework, we can deduce that the FL model is decentralized since each blockchain node has the opportunity to engage in local model training and contribute to global model aggregation. Thus, the blockchain effectively assumes the role of a central aggregator.

Within this setup, there are two approaches for averaging the global model: (a) Selected nodes collect validated local model updates and execute the aggregation algorithm. (b) All nodes actively participate in the global model aggregation process.

The distributed ledger records the training data, including validated local model updates, global model updates, and other pertinent data generated during the learning process. The general workflow of BCFL can be outlined as follows:

\begin{figure}[h]
\centering
\includegraphics[width=0.7\textwidth]{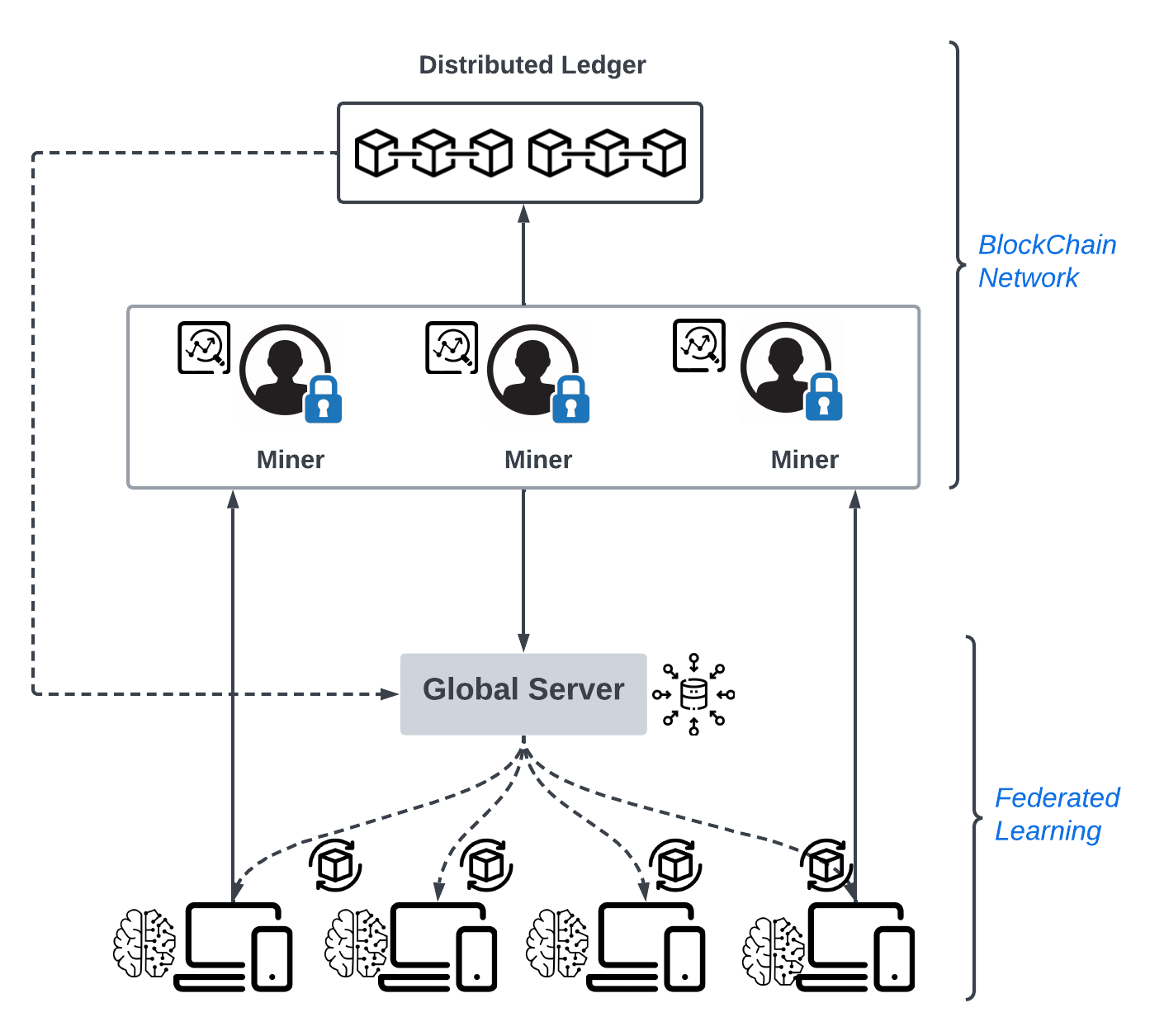}
\caption{This architecture leverages the strengths of both federated learning and blockchain technology, enabling secure, efficient, and privacy-preserving distributed machine learning. The decentralized nature of blockchain ensures that there is no single point of failure, and the federated learning approach maintains data privacy by keeping data localized.} 
\label{fig:blockchain}
\vspace{-10pt}
\end{figure}

The clients are tasked with collecting data and training models on their local devices. The local model updates are then verified by selected clients. After verification, these updates are gathered by chosen clients to update the global model. A new block, containing the verified model updates, is subsequently added to the distributed ledger. In line with the incentive mechanism, rewards are distributed to the participants.

BCFL has been explored in various research studies. In one study~\cite{weng2019deepchain}, FL clients are positioned at the edge, where they can collect sensor data and provide computational power. These clients handle data collection and training. The blockchain acts as a distributed ledger in this framework, recording the training data. This setup ensures the integrity of the raw data and prevents malicious clients from undermining the process.

In another study~\cite{hua2020blockchain}, a blockchain-based FL system was proposed, where participants competed to generate new blocks. The winner would gather the model parameters and update the blockchain. Since no raw data was shared during the training process, this system effectively preserved data privacy securely.

In~\cite{toyoda2019mechanism}, a blockchain-integrated FL platform was designed, assuming that all participants would behave rationally under a competition incentive mechanism. This platform could handle various types of raw data such as text, audio, and images. Before uploading local model updates, a security procedure was conducted by selected workers under the smart contract to ensure data validity.

BAFELE, a decentralized FL framework using blockchain, was introduced as a solution without a central aggregator~\cite{ramanan2020baffle}. By organizing the FL process into rounds and collecting local model updates to update the global model, BAFELE achieved comparable model training performance to conventional FL models while using fewer computational resources.

\subsection{Adaptive Federated Learning}

When dealing with a substantial volume of data required to train a precise model, the federated learning procedure can place a considerable burden on resources. The term "resources" encompasses various factors such as time, energy, monetary expenses, and both computational and communication aspects. It is often necessary to impose restrictions on the resources utilized for model training to prevent system congestion and maintain cost-effectiveness. This becomes especially crucial in edge computing settings where computational and communication resources are not as plentiful as those found in data centers.

\begin{figure}[h]
\centering
\includegraphics[width=0.7\textwidth]{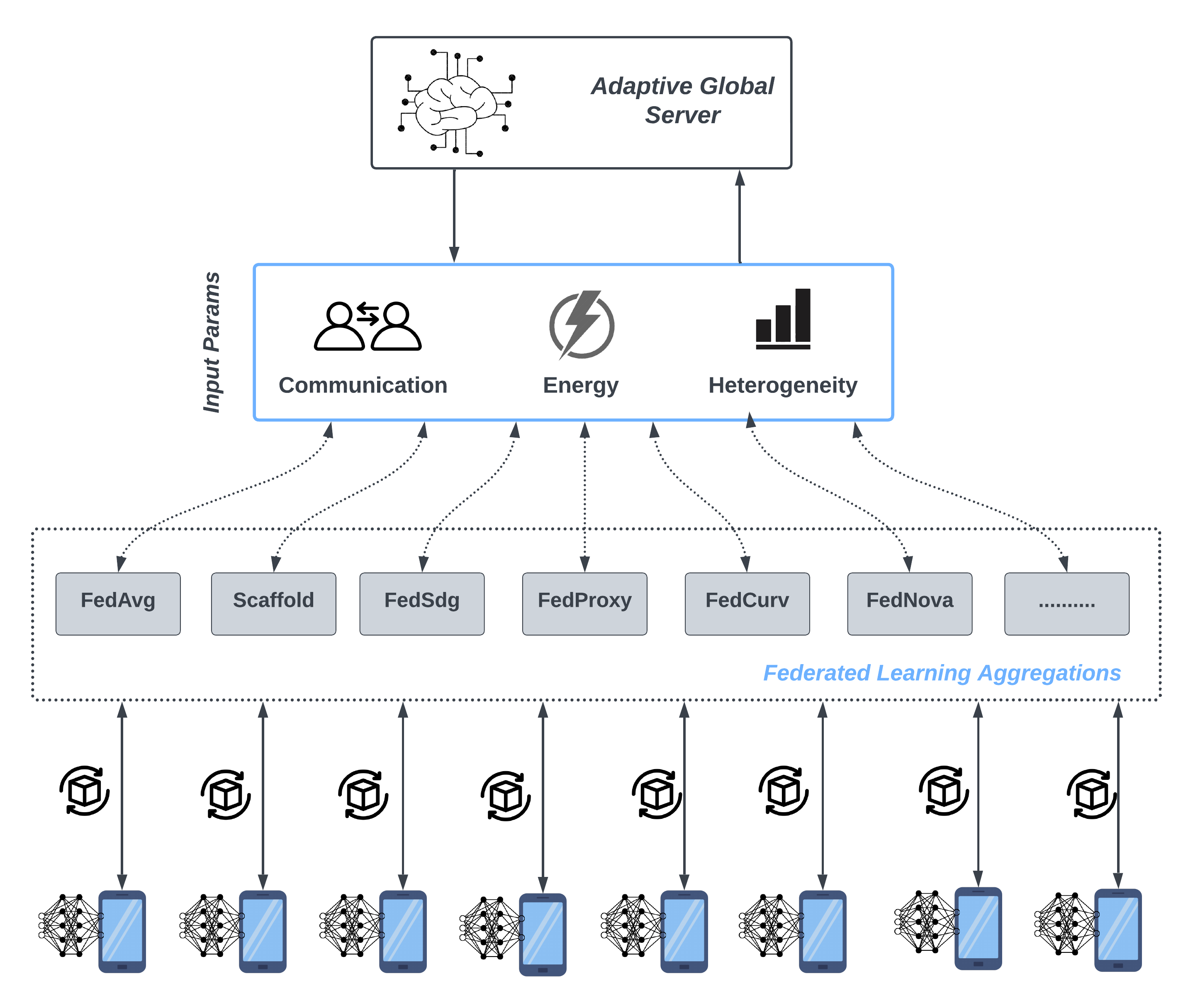}
\caption{This architecture demonstrates the integration of efficient fine-tuning parameters within federated learning, focusing on optimizing communication, conserving energy, and addressing data and device heterogeneity. These optimizations ensure that federated learning can be effectively deployed in real-world scenarios with diverse and resource-constrained environments.} 
\label{fig:adaptive-fl}
\vspace{-10pt}
\end{figure}

\section{Generative AI}
Generative AI is a branch of artificial intelligence focused on developing models and algorithms that can create new content resembling data from a particular domain. The fundamental concept of generative AI involves training models to understand the underlying patterns and structures of a dataset, enabling them to generate new samples with similar characteristics.

Generative models seek to capture the probability distribution of the training data and produce new instances by sampling from this learned distribution. These models can be trained using either supervised or unsupervised methods, depending on whether labeled data is available.

Several types of generative models are commonly used in generative AI, including:

\begin{itemize}
  \item \textit{Generative Adversarial Networks (GANs):} GANs comprise two neural networks: a generator and a discriminator. The generator creates new samples, while the discriminator attempts to distinguish between real and generated samples. Through an adversarial training process, the generator improves its ability to produce samples that can deceive the discriminator, resulting in the generation of realistic and high-quality content.
  \item \textit{Variational Autoencoders (VAEs):} VAEs are probabilistic models consisting of an encoder and a decoder. The encoder maps input data to a latent space representation, and the decoder reconstructs the input data from this latent space. By sampling from the latent space, VAEs can generate new samples that mimic the training data distribution.
  \item \textit{Autoregressive Models:} Autoregressive models estimate the conditional probability distribution of a sequence of variables by modeling the dependencies between each variable and its predecessors. These models generate new samples by iteratively predicting each subsequent variable based on the preceding ones.
  \item \textit{Flow-Based Models:} Flow-based models learn a series of invertible transformations that map a simple distribution (e.g., Gaussian) to a more complex distribution resembling the training data distribution. These models facilitate efficient sampling and exact likelihood computation.
\end{itemize}


Generative AI has been applied across diverse fields, including image synthesis, text generation, music composition, video creation, and more. It has facilitated tasks such as image-to-image translation, style transfer, data augmentation, anomaly detection, and the development of recommender systems.

The progress in generative AI has been fueled by the availability of large-scale datasets, enhanced computational resources, and advancements in deep learning techniques. Ongoing research continues to explore and develop new generative models and methods to improve the quality, diversity, and controllability of generated content.

Integrating generative AI techniques into federated learning offers numerous opportunities to boost privacy, data efficiency, and model performance. We provide an overview of these advancements, highlighting the potential of generative AI to transform federated learning and address its challenges. However, it is crucial to consider the ethical implications, potential biases, and limitations when implementing generative AI in federated learning scenarios.

\subsection{Use Cases for Generative AI}
\subsubsection{Text Generation}
Text generation using generative AI in federated learning involves leveraging generative models within the context of federated learning to generate text that aligns with the data distribution of decentralized participants without sharing their raw data. This approach enables the generation of text samples that capture the characteristics and patterns of the distributed data sources while maintaining data privacy and security.

\subsubsection{Text Summarization}
Text summarization using generative AI involves the use of artificial intelligence algorithms, particularly natural language processing (NLP) techniques, to automatically generate concise summaries of longer texts. Generative AI models, such as the GPT-3.5 model, can be utilized to accomplish this task.

\subsubsection{Text Extraction}

Text extraction with generative AI involves the automatic retrieval of specific information or data from text using generative artificial intelligence models. This technique trains models to recognize and extract pertinent entities, relationships, or structured data from unstructured text.

This method of text extraction can be advantageous for various applications, including information retrieval, data analysis, and knowledge extraction from extensive text corpora. However, the accuracy of the extraction process is highly dependent on the quality of the training data, the design of the model, and the complexity of the extraction task. Regular evaluation and fine-tuning of the model are often necessary to enhance extraction performance.

\subsubsection{Paraphrase Rephrase}

Paraphrasing or rephrasing with generative AI involves creating alternative versions of a given text while maintaining its original meaning. Generative AI models, like GPT-3.5, can be employed to automatically produce paraphrases by utilizing their language comprehension and generation capabilities.

However, it is important to recognize that the quality and accuracy of these paraphrases can vary. The generated paraphrases might not always be perfect or contextually suitable. Therefore, human review and editing are often necessary to ensure that the paraphrases are both high-quality and relevant, especially in sensitive or complex areas.

Furthermore, caution should be exercised when using generative AI for paraphrasing due to potential ethical considerations, such as avoiding plagiarism and respecting copyright laws. Generative AI should be used as an assistive tool in the paraphrasing process, with all generated paraphrases reviewed and validated before being used in any critical context.

\subsubsection{Smart Search Tool}

A smart search tool using generative AI refers to a system that utilizes artificial intelligence algorithms, particularly generative models, to enhance the search experience and provide more intelligent and relevant search results to users. This tool leverages the power of generative AI models, such as GPT-3.5, to understand and interpret user queries, generate more accurate search results, and improve the overall search experience.

A smart search tool using generative AI has the potential to enhance search experiences by providing more intelligent, context-aware, and personalized results. However, it's important to ensure transparency, accountability, and user privacy when employing generative AI in search applications. Regular monitoring and human review are crucial to maintain the quality and fairness of the search results generated by the AI system.

\subsubsection{Image Generation}

Image generation with generative AI involves creating new and realistic images through artificial intelligence algorithms, specifically generative models. Techniques such as generative adversarial networks (GANs) and variational autoencoders (VAEs) are commonly used to produce visually appealing and coherent images by learning patterns and structures from a training dataset.

It is important to recognize that the capabilities and quality of output from generative AI models for image generation can vary. Advanced models like StyleGAN or BigGAN have demonstrated remarkable results in creating highly realistic and diverse images. However, it is crucial to remember that these generated images are based on patterns learned from the training data and may not always accurately depict real-world objects or scenes.

Moreover, ethical considerations and responsible use are essential when generating images with generative AI. Care must be taken to prevent the use of generated images for malicious purposes, such as creating fake or misleading content.

In summary, image generation using generative AI holds significant potential for various applications, including art, design, content creation, and data augmentation in machine learning tasks.

\subsubsection{Chat bot}

A chatbot powered by generative AI is an AI-driven conversational agent that leverages generative models, such as GPT-3.5 or similar architectures, to comprehend and respond to user queries in a conversational style. These chatbots are designed to engage in natural language interactions, delivering relevant and contextually appropriate replies.

While generative AI-based chatbots can produce coherent and contextually relevant responses, they may sometimes generate inaccurate or nonsensical answers. The quality of these responses is significantly influenced by the training data, the model's design, and the complexity of the conversational task.

It is also crucial to implement human oversight and moderation when deploying chatbots to ensure they provide accurate and appropriate responses. Continuous evaluation, monitoring, and iterative improvements are vital to enhance the chatbot's performance and address potential limitations or biases in its replies.

In summary, chatbots utilizing generative AI can be applied in various domains, such as customer support, virtual assistants, or interactive conversational agents, to offer users human-like interactions and assistance.

\section{Integrating generative models and ml models using federated learning}

\vspace{-.25cm}

\begin{algorithm}
  \caption{\textbf{Federated Averaging Algorithm}}
  \label{alg:fed_avg}
  \textbf{Input:} $S_{T}$, $K$, policy $P$, $p_{k}(i) \triangle k, i$ \\
  \textbf{Initialize:}
  \begin{itemize}
    \item Set hyperparameters $hp$
    \item Split data: $train, test \leftarrow \text{split}(X)$
    \item Split data further: ${C}_{x} \leftarrow \text{split}(train, test)$
    \item Initialize global model $w^o$
  \end{itemize}
  \begin{algorithmic}[1]
    \For{$t = 1, 2, \ldots, T$}
      \For{$k \in {C}_{x}$}
        \State Send $w^{t-1}$ to client
        \State $w^{t}_{k} \leftarrow \text{CLIENTUPDATE}(hp, m, w^{t-1})$
      \EndFor
      \State ${w}^{t} \leftarrow {w}^{t-1} + \frac{\eta}{N} \sum_{k=1}^{N}({w}^{t}_{k} - {w}^{t-1})$
    \EndFor
    \State $w^T$ represents the optimal weight after $T$ rounds
  \end{algorithmic}
  \textbf{Return:} model accuracy and precision
\end{algorithm}

\section{Applications of generative AI models in Federated Learning}

By integrating generative AI tools like ChatGPT~\cite{radford2019language}, users can experience enhanced benefits. For instance, in a smart home setting, the synergy between generative artificial intelligence and Federated Learning can be employed to identify unfamiliar individuals approaching the property and convert that information into a courteous notification for the homeowner. To exemplify, if an ex-convict or an individual with a criminal history is detected near the home, the Federated model would identify the person and issue an alert. The generative AI component would then transform this alert into a polite message, informing the homeowner about the stranger's identity.

\subsection{Smart Homes}

The goal of integrating the Federated Learning (FL) approach within a smart home environment is to enable smart devices to progressively improve their individual models, which can vary across devices. An example scenario involves the periodic transmission of these models to the cloud platforms of their respective manufacturers. As a result, the shared generic model receives updates along with updates from other users who own the same device model but live in different smart homes. Once the shared generic model is updated, it can be redistributed to each smart device, much like how text prediction models are deployed on smartphones~\cite{mcmahan2017communication}. This process benefits not only the owner of the smart device, whose device improves based on their behavior, but also other users with the same smart device.

For instance, consider a situation where a smart doorbell with facial recognition capabilities detects an unfamiliar person approaching the homeowner's residence. The Federated Learning model integrated into the smart home system identifies the individual's presence and generates an alert. At the same time, the generative AI component converts this alert into a courteous audio message, informing the homeowner about the visitor's identity and providing relevant details based on the facial recognition data. This proactive method enhances both security and convenience for the homeowner, enabling them to make informed decisions about granting access to their home~\cite{chen2020smart}.

\begin{figure}[h]
\centering
\includegraphics[width=0.7\textwidth]{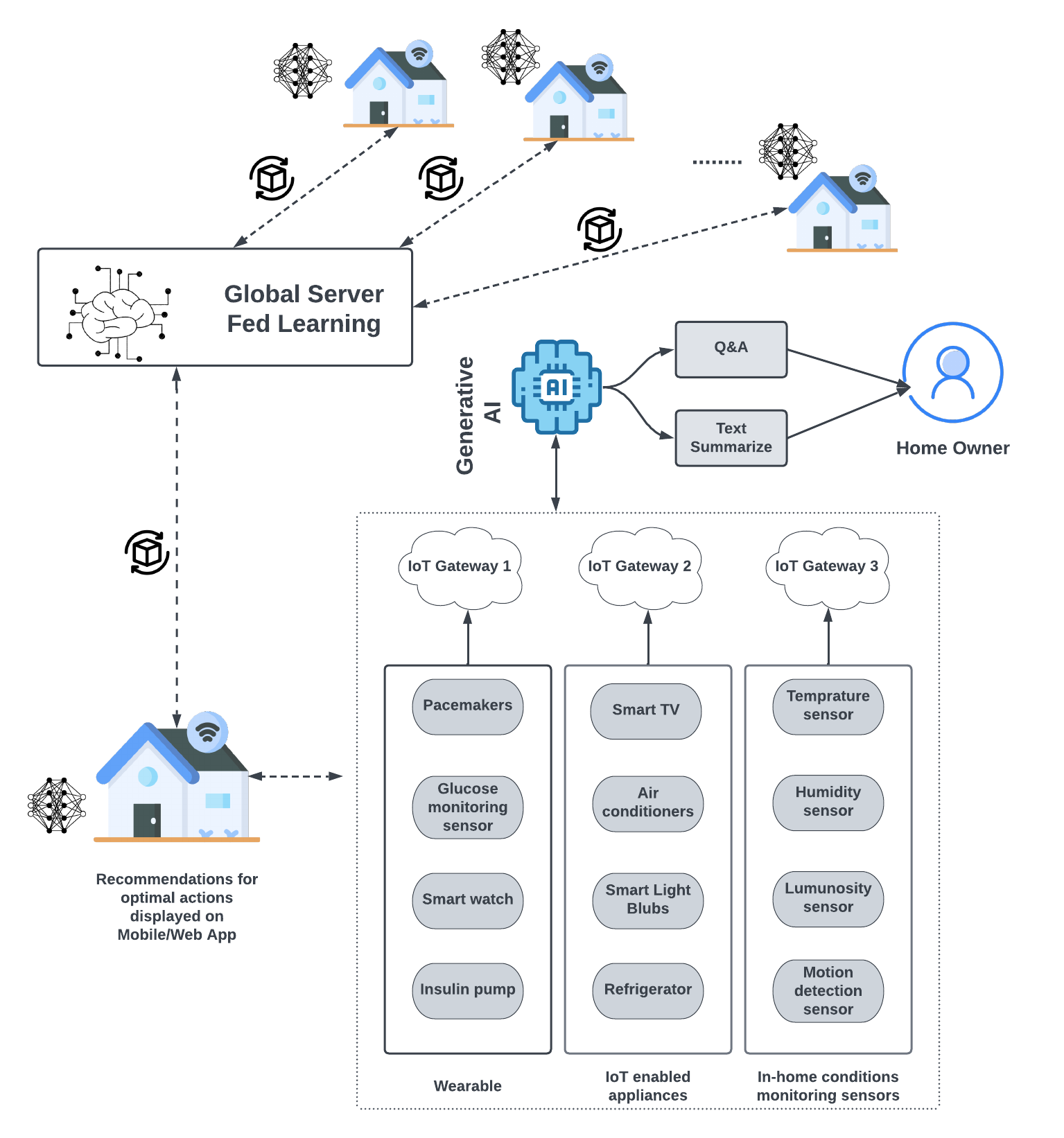}
\caption{By leveraging federated learning and generative AI, smart home devices can provide advanced, personalized services while preserving user privacy and enhancing security. The integration of blockchain technology further ensures the integrity and transparency of the learning process, making it a robust solution for modern smart home environments.} 
\label{fig:smart-homes}
\vspace{-10pt}
\end{figure}

However, utilizing the federated learning approach in this particular scenario introduces additional network costs. Specifically, when an update is initiated, access to an external network is required. Subsequently, the manufacturer sends an update of the shared generic model to the network, which might necessitate user consent before implementing the update on the device. Moreover, the updates need to be exchanged in both directions (i.e., between smart devices and clouds), resulting in increased overall costs and potentially occurring more frequently than conventional updates.

\subsection{Health Care}

Federated Learning (FL) holds significant promise in the healthcare sector as a groundbreaking approach for safeguarding data privacy. While individual medical institutions possess substantial amounts of patient data, it may not be sufficient to train accurate prediction models on their own~\cite{szegedi2019federated}. The integration of FL with disease prediction serves as a viable solution to overcome the challenges associated with analyzing data across various hospitals. By utilizing FL, barriers to collaboration in data analysis can be dismantled, enabling enhanced prediction models that draw insights from multiple healthcare facilities.

Consider a scenario where multiple hospitals collaborate to improve the accuracy of a disease prediction model using Federated Learning (FL). Each hospital has access to its patient data, but due to privacy regulations and concerns, sharing this data directly is not feasible. With FL, instead of sharing raw data, hospitals train local models on their data and only share model updates with a centralized server. The server aggregates these updates and computes a global model, which is then sent back to each hospital for further refinement. This collaborative approach allows hospitals to collectively improve the accuracy of the disease prediction model while ensuring patient privacy and data security~\cite{smith2020federated}.

\begin{figure}[h]
\centering
\includegraphics[width=0.7\textwidth]{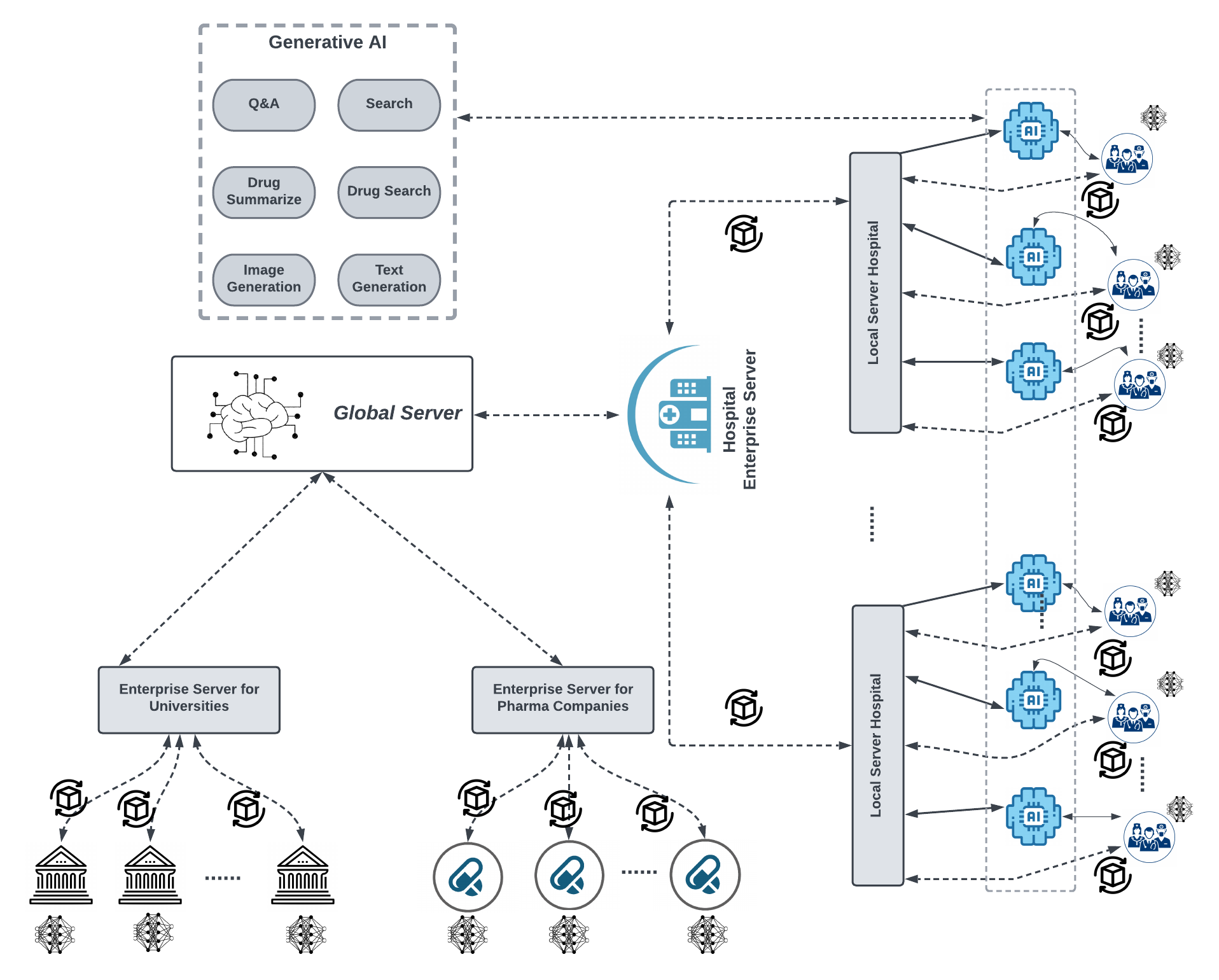}
\caption{By leveraging federated learning and generative AI, healthcare providers can collaboratively improve medical models and treatment strategies while preserving patient privacy. The integration of blockchain technology further ensures the integrity and transparency of the learning process, making it a robust solution for advancing healthcare through secure and efficient data sharing and analysis.} 
\label{fig:hospital}
\vspace{-10pt}
\end{figure}

\subsubsection{Personalized Treatment Recommendations}

Federated learning can be employed to train generative AI models on patient health data from multiple sources. This can enable the development of personalized treatment recommendation systems. Each healthcare provider can contribute their data, and the model can learn from diverse patient populations to offer tailored treatment suggestions without compromising individual patient privacy.

\subsubsection{Anomaly Detection and Early Disease Diagnosis}

Generative AI models, trained using federated learning, can be utilized to detect anomalies or early signs of diseases across different healthcare settings. By collectively training on distributed data, these models can learn patterns and recognize abnormalities, aiding in early diagnosis and intervention.

\subsubsection{Drug Discovery and Development}

Federated learning enables collaboration between pharmaceutical companies, research institutions, and healthcare providers for drug discovery. Generative AI models can assist in generating new molecules, predicting drug interactions, and optimizing drug design. By leveraging federated learning, organizations can pool their knowledge and expertise without sharing proprietary data.

\subsubsection{Natural Language Processing for Clinical Documentation}

Generative AI models trained on federated data can enhance natural language processing capabilities for clinical documentation. These models can improve tasks like automated summarization of patient records, extracting relevant information, and generating accurate and concise medical reports.

\subsection{IOT Devices}

\subsubsection{Edge-based Anomaly Detection}
 IoT devices generate a vast amount of sensor data, and detecting anomalies in real-time is crucial for maintaining device functionality and security. By deploying generative AI models trained with federated learning directly on edge devices, anomalies can be detected and flagged locally, reducing the need for constant data transmission to centralized servers.

 \subsubsection{Privacy-Preserving Smart Surveillance}
 Federated learning enables collaborative training of generative AI models for smart surveillance systems. Instead of sending sensitive video streams to a central server for analysis, local devices can train and refine models using federated learning. This preserves privacy by keeping the data local, while still benefiting from improved video analysis capabilities.

 \subsubsection{Adaptive User Interfaces}
 Generative AI models trained using federated learning can enhance user interfaces on IoT devices. These models can learn from diverse user interactions across different devices, leading to personalized and adaptive interfaces that cater to individual preferences and behaviors.

 \subsubsection{Context-Aware Recommender Systems}

 Federated learning enables IoT devices to collaboratively train generative AI models for context-aware recommendations. For example, a network of smart home devices can share information about user preferences and behaviors to collectively train a model that provides personalized recommendations for music, lighting, or other smart home settings.

 \subsubsection{Predictive Maintenance and Fault Detection}

 Generative AI models trained using federated learning can analyze sensor data from IoT devices to predict maintenance needs and detect potential faults. By sharing insights and collectively learning from distributed data, these models can identify patterns and anomalies, enabling proactive maintenance and reducing downtime.
 
\section{Future Scope}
The integration of generative AI with blockchain federated learning (BCFL) opens up several exciting avenues for future research and practical applications. One of the primary areas of focus will be the enhancement of privacy-preserving techniques. As privacy concerns continue to rise, it will be crucial to develop generative models that can produce synthetic data indistinguishable from real data while ensuring that no sensitive information is leaked. Future research could explore advanced differential privacy techniques and secure multi-party computation to further bolster the privacy guarantees of BCFL systems. This will be especially important in sectors like healthcare, where data sensitivity is paramount.

Scalability is another critical aspect that requires attention. As the number of devices and participants in federated learning networks grows, ensuring efficient and scalable model training and updating processes will be essential. Researchers will need to develop algorithms and architectures that can handle the complexities of large-scale deployments, including optimizing communication overhead and computational resource allocation. Innovative solutions such as hierarchical federated learning, where intermediate aggregations occur at different levels before a final global aggregation, could be explored to alleviate scalability issues.

The potential for real-time applications of BCFL is vast, particularly in areas requiring instantaneous data processing and decision-making. For instance, in smart surveillance systems, real-time federated learning could enable immediate identification and response to security threats. Similarly, in healthcare, real-time data from wearable devices could be used to continuously update and refine predictive models for patient monitoring and diagnosis. Developing low-latency communication protocols and efficient real-time data processing algorithms will be crucial to realizing these applications.

Edge computing integration presents another promising area for future development. By leveraging edge devices to perform local computations, BCFL can reduce the latency and bandwidth requirements associated with centralized cloud processing. This approach will be particularly beneficial for IoT applications, where devices often operate in resource-constrained environments. Future research could focus on optimizing edge-based federated learning frameworks, ensuring they can operate efficiently with limited computational power and intermittent connectivity.

Ethical considerations and bias mitigation will also be essential components of future BCFL research. As generative models are integrated into federated learning systems, ensuring that these models do not perpetuate or amplify existing biases will be crucial. Researchers will need to develop methods for detecting and mitigating bias in generative models, ensuring that the outcomes are fair and equitable. Additionally, transparent reporting and accountability mechanisms will be necessary to address ethical concerns and build trust in BCFL systems.

Finally, establishing interoperability standards will be vital for the broader adoption of BCFL. Developing industry-wide protocols and standards will facilitate seamless collaboration and data sharing across different platforms and organizations. This will be particularly important in sectors like healthcare and finance, where data from diverse sources needs to be integrated to build comprehensive predictive models. Standardization efforts could include defining common data formats, communication protocols, and security measures to ensure compatibility and interoperability between different BCFL systems.

By addressing these key areas, future research and development in generative AI and blockchain federated learning can unlock the full potential of this innovative approach, driving advancements across various industries and enhancing the overall efficacy and security of distributed machine learning systems.

\section{Conclusions}
The integration of generative AI with blockchain federated learning represents a promising advancement in the field of distributed machine learning. By leveraging the strengths of generative models, such as synthetic data generation and privacy preservation, BCFL can address many of the challenges associated with traditional federated learning. The incorporation of blockchain technology further enhances the security and decentralization of the learning process. Practical applications in healthcare, smart homes, and IoT devices showcase the potential of this approach to revolutionize various industries by providing more secure, efficient, and personalized solutions. However, further research is needed to fully realize the potential of this technology, addressing issues related to scalability, real-time processing, and ethical considerations.


\end{document}